\documentclass[runningheads]{llncs}
\usepackage{amssymb}
\usepackage{cite}
\usepackage{graphicx}
\usepackage{amsmath}

\usepackage{graphicx}
\usepackage{booktabs}

\usepackage{array}
\usepackage{adjustbox}
\usepackage{relsize}

\usepackage{multirow}

\begin{document}
%
\title{DRAM Failure Prediction in AIOps: Empirical Evaluation, Challenges and Opportunities}

\author{Zhiyue Wu\inst{1,2}, Hongzuo Xu\inst{1,2}, Guansong Pang\inst{3}, Fengyuan Yu\inst{2}, Yijie Wang\inst{1,2}, Songlei Jian\inst{2}, and Yongjun Wang\inst{2}}

\authorrunning{Z. Wu et al.}
\titlerunning{DRAM Failure Prediction in AIOps}

\institute{
$^1$ Science and Technology on Parallel and Distributed Processing Laboratory \\
$^2$ College of Computer, National University of Defense Technology, China \\
$^3$ University of Adelaide, Australia \\
\email{zhiyuewu@126.com}
\email{\{xuhongzuo13,wangyijie,jiansonglei,wangyongjun\}@nudt.edu.cn} \\
\email{guansong.pang@adelaide.edu.au}}

\maketitle

\begin{abstract}
DRAM failure prediction is a vital task in AIOps, which is crucial to maintain the reliability and sustainable service of large-scale data centers. 
However, limited work has been done on DRAM failure prediction mainly due to the lack of public available datasets. This paper presents a comprehensive empirical evaluation of diverse machine learning techniques for DRAM failure prediction using a large-scale multi-source dataset, including more than three millions of records of kernel, address, and mcelog data, provided by Alibaba Cloud through PAKDD 2021 competition.
Particularly, we first formulate the problem as a multi-class classification task and exhaustively evaluate seven popular/state-of-the-art classifiers on both the individual and multiple data sources. We then formulate the problem as an unsupervised anomaly detection task and evaluate three state-of-the-art anomaly detectors. Further, based on the empirical results and our experience of attending this competition, we discuss major challenges and present future research opportunities in this task.

\end{abstract}

\section{Introduction}
The past decade has witnessed great development of cloud services \cite{wang2017joint}. Large-scale data center is infrastructure of cloud computing, which provides essential support to upper cloud applications. In production data centers, cloud providers suffer from frequent occurrence of hardware failures, in which Dynamic Random Access Memory (DRAM) failure is one of the main causes of hardware failures \cite{wang2017what}. These failures may cause severe outages, which leads to large economic costs and violates the business agreements with the users. Therefore, both of the academic and industry communities show increasing interests in exploring techniques to ensure the reliability of large-scale data centers. 


Limited work has been done
on predicting DRAM failures due to the lack of public available datasets. Literature \cite{Hwang2012Cosmic,Schroeder2011draw} provides analytical studies on DRAM errors and their characteristics, but they do not show how to predict DRAM errors. In recent years, a number of methods \cite{Giurgiu2017Predicting,boixaderas2020cost,Patwari2017Understanding,Du2018Memory, Mukhanov2019Workload, Baseman2018Physics} are proposed to predict DRAM failures. 
However, to the best of our knowledge, no public available dataset is released for the task, as the DRAM failure data is highly sensitive. This significantly hinders the development and evaluation of machine learning techniques to address the DRAM failure prediction problem.


To promote the development of this research line, Alibaba Cloud holds a PAKDD 2021 competition on DRAM failure prediction using large-scale multi-source data extracted from its own production data centers\footnote{https://tianchi.aliyun.com/competition/entrance/531874/}. We submit a solution to this competition. Up to now, our solution obtains 32.64 score online in the second round of the competition. This paper is based on our exploration on the six-month data of DRAM failures of large-scale data centers in this competition. The dataset contains more than three millions of records of kernel, address, and mcelog data.

Particularly, this paper presents a comprehensive empirical evaluation of diverse machine learning techniques for DRAM failure prediction using the large-scale multi-source dataset described above. We first formulate the problem as a multi-class classification task and exhaustively evaluate seven popular/state-of-the-art classifiers on both the individual and multiple data sources. We then formulate the problem as an unsupervised anomaly detection task and evaluate three state-of-the-art anomaly detectors. Further, based on the empirical results and our experience of attending this competition, we discuss major challenges and present future research opportunities in this task.


\section{Related Work}


In large-scale data centers, a single job may execute for days on thousands of nodes. All CPU can waste for long time if DRAM failure occurs at any of these nodes \cite{boixaderas2020cost}. Additionally, the advancing densities in DRAM may lead to more failures. It is highly demanding to reduce the costs due to hardware replacement and service disruption by accurately predicting DRAM failures. 

There have been a few recent studies on addressing this problem. \cite{Giurgiu2017Predicting} uses ensemble learning techniques to predict uncorrectable errors weeks in advance. A spatial analysis framework \cite{Patwari2017Understanding} is developed to visualize the failures. Based on the historical memory failure data, an online learning method \cite{Du2018Memory} is proposed to perform the prediction by modeling some implicit patterns. Three machine learning models (SVM, KNN, and Random Forest) are used to predict DRAM failure in a single server in \cite{Mukhanov2019Workload}.


Cost-benefit calculation is used in \cite{jauk2019sc,boixaderas2020cost} to yield new metrics apart from commonly-used classification evaluation methods (e.g., Precision, Recall, and F1-Score). These new metrics are more suitable to determine whether prediction is useful in practice.

Due to the imbalance nature of positive and negative samples, data noise, and other uncertainties, the above methods may fail to obtain desired results in large-scale production data centers. 



\section{Empirical Evaluation}
\subsection{Experimental Setups}


\subsubsection{Datasets}
We first pre-process the raw data, perform feature engineering, and then use a labeling method based on the given failure table to yield class labels. Our source code will be released after the competition.


Specifically, we generate features from three given tables and yield three new datasets. Table \ref{tab:basic} reports basic characteristics of datasets generated from the competition raw data. We show the number of generated features, the number of records, and the number of positive/negative records. As we set three different positive classes according to their time interval to real server failures, the number of three different classes is also reported. 

These tables can be further merged by the collected time of records to further generate new combined datasets. In our solution, two combined tables are used, i.e., Address\_Mce and Kernel\_Address. Please note that the number of records in Kernel\_Address is large than the Kernel table. This is because some records in the Kernel table correspond to multiple records in the Address table.  


\begin{table}[htbp]
  \centering
  \caption{Basic Characteristics of Datasets.}
  \scalebox{0.8}{
    \begin{tabular}{cccccc}
    \toprule
    \multicolumn{1}{c}{\textbf{Data}} & \multicolumn{1}{c}{\textbf{\#features}} & \multicolumn{1}{c}{\textbf{\#records}} & \multicolumn{1}{c}{\textbf{\#negatives}} & \multicolumn{1}{c}{\textbf{\#positives}} & \multicolumn{1}{c}{\textbf{\#positive classes}} \\
    
    \midrule
    Kernel & 55    & 650,323 & 649,074 & 1,294 & 455/286/508 \\
    Address & 19    & 1,451,013 & 1,449,687 & 1,326 & 370/321/635 \\
    Mce   & 82    & 1,483,197 & 1,481,869 & 1,328 & 370/321/637 \\
    \hline
    Address\_Mce & 99    & 1,451,013 & 1,449,687 & 1,326 & 370/321/635 \\
    Kernel\_Address & 72    & 650,653 & 649,403 & 1,250 & 455/287/508 \\
    \bottomrule
    \end{tabular}}
  \label{tab:basic}%
\end{table}%

\subsubsection{Evaluation Measures}
As we have a training set spanning six months, we use a six-fold cross-validation method to train and validate the performance of testing classifiers. Specifically, the full training set is partitioned into 6 subsets w.r.t. the collected month of these records. One subset is treated as validation set, and the union of the remaining five subsets is used as training set. This process is repeated six times (the folds). Models can be validated in the data of each month. The six results can then be combined to produce the final estimation. All the data in the full training set can be employed for both training and validation. 

\begin{table}[ht!]
  \centering
  \caption{Six-fold Cross-validation Results ($P$/$R$/$F_1$) of Different Classifiers on Five Datasets. The classifier that obtains the best $F_1$ per dataset per fold is boldfaced. }
    \scalebox{0.75}{
    \begin{tabular}{p{1.1cm}<{\centering}|p{2.4cm}<{\centering}|p{2.4cm}<{\centering}|p{2.4cm}<{\centering}|p{2.4cm}<{\centering}|p{2.4cm}<{\centering}|p{2.4cm}<{\centering}}
    \toprule
     \multicolumn{1}{c}{}    & \multicolumn{1}{c}{\textbf{Jan-Fold}} & \multicolumn{1}{c}{\textbf{Feb-Fold}} & \multicolumn{1}{c}{\textbf{\textbf{Mar-Fold}}} & \multicolumn{1}{c}{\textbf{Apr-Fold}} & \multicolumn{1}{c}{\textbf{May-Fold}} & \multicolumn{1}{c}{\textbf{Jul-Fold}} \\
    \midrule
    \multicolumn{7}{c}{\textbf{Kernel}}\\
    \hline
    SVM   &  65.8/27.8/39.1 &  74.5/22.2/34.1 &  70.0/21.5/32.9 &  76.5/16.0/26.5 &  65.2/17.3/27.4 &  52.0/21.8/30.8 \\
    LR    &  67.2/23.9/35.2 &  72.7/20.3/31.7 &  69.6/19.6/30.6 &  75.9/13.6/23.0 &  60.4/18.5/28.3 &  25.6/17.6/20.9 \\
    RF    &  34.1/33.3/33.7 &  43.6/36.7/39.9 &  46.4/35.6/40.3 &  37.9/29.0/32.9 &  32.1/26.0/28.8 &  26.7/32.8/29.4 \\
    DT    &  17.5/40.6/24.5 &  21.5/46.2/29.4 &  22.3/48.5/30.5 &  18.0/40.7/25.0 &  18.7/42.2/25.9 &  11.8/42.9/18.4 \\
    GBDT  &  \textbf{60.9/31.1/41.2} &  63.2/30.4/41.0 &  50.6/25.2/33.6 &  59.6/19.1/29.0 &  51.5/20.2/29.0 &  45.3/24.4/31.7 \\
    XGB   &  42.1/32.8/36.9 &  \textbf{47.0/39.9/43.2} &  48.9/39.3/43.5 &  41.3/30.9/35.3 &  37.1/30.1/33.2 &  27.3/35.3/30.8 \\
    LGBM   &  50.0/33.3/40.0 &  50.0/34.2/40.6 &  \textbf{55.0/37.4/44.5} &  \textbf{50.0/33.3/40.0} & \textbf{39.4/30.1/34.1} &  \textbf{29.7/34.5/31.9} \\
        
    \hline
    \multicolumn{7}{c}{\textbf{Address}}\\
    
    \hline
    
        SVM   &  25.0/0.22/0.41 &  42.1/0.51/0.90 &  52.4/0.67/12.0 &  75.0/0.37/0.71 &  45.5/0.58/10.3 &  34.4/0.92/14.6 \\
    LR    &  33.3/0.06/0.11 &  0.00/0.00/0.00 &  33.3/0.12/0.24 &  0.00/0.00/0.00 &  0.00/0.00/0.00 &  42.9/0.25/0.48 \\
    RF    &  30.0/15.0/20.0 &  35.0/17.7/23.5 &  37.6/21.5/27.3 &  35.4/21.6/26.8 &  35.3/17.3/23.3 &  20.2/16.8/18.3 \\
    DT    &  16.1/36.7/22.3 &  14.9/38.0/21.4 &  18.8/45.4/26.6 &  17.1/43.8/24.7 &  15.0/37.0/21.4 &  0.81/32.8/13.0 \\
    GBDT  &  23.5/17.8/20.3 &  30.0/22.8/25.9 &  38.8/16.0/22.6 &  35.1/16.7/22.6 &  32.8/23.7/27.5 &  20.4/16.0/17.9 \\
    XGB   &  \textbf{33.9/21.1/26.0} &  34.0/20.3/25.4 &  35.1/20.9/26.2 &  \textbf{35.9/22.8/27.9} &  31.3/20.8/25.0 &  17.8/17.6/17.7 \\
    LGBM   &  20.4/28.9/23.9 &  \textbf{28.9/27.8/28.4} &  \textbf{28.7/28.8/28.7} &  26.3/25.3/25.8 &  \textbf{27.9/37.0/31.8} &  \textbf{16.1/26.1/19.9} \\

    \hline
    \multicolumn{7}{c}{\textbf{Mce}}\\
    \hline
    
        SVM   &  62.5/0.28/0.53 &  66.7/0.51/0.94 &  71.4/0.61/11.3 &  77.8/0.43/0.82 &  40.0/0.35/0.64 &  60.0/0.76/13.4 \\
    LR    &  25.0/0.17/0.31 &  0.00/0.00/0.00 &  50.0/0.12/0.24 &  22.2/0.12/0.23 &  16.7/0.06/0.11 &  25.0/0.17/0.31 \\
    RF    &  18.0/21.7/19.6 &  18.7/29.7/22.9 &  22.1/34.4/26.9 &  17.4/28.4/21.6 &  16.6/28.9/21.1 &  12.4/24.4/16.5 \\
    DT    &  \textbf{15.4/32.8/20.9} &  16.4/36.1/22.5 &  16.8/40.5/23.7 &  14.9/35.2/21.0 &  13.5/35.8/19.6 &  10.9/31.1/16.2 \\
    GBDT  &  28.2/11.1/15.9 &  25.7/11.4/15.8 &  37.5/14.7/21.1 &  31.5/10.5/15.7 &  21.8/11.0/14.6 &  29.3/14.3/19.2 \\
    XGB   &  22.1/15.0/17.9 &  27.8/23.4/25.4 &  31.2/23.9/27.1 &  \textbf{29.8/22.8/25.9} &  26.7/23.1/24.8 &  18.0/19.3/18.6 \\
    LGBM   &  22.3/19.4/20.8 &  \textbf{24.7/29.7/27.0} &  \textbf{28.5/28.8/28.7} &  25.2/25.3/25.2 & \textbf{25.0/25.4/25.2} &  \textbf{18.2/26.1/21.5} \\
    
    \hline
    \multicolumn{7}{c}{\textbf{Address\_Mce}}\\
    \hline
    
    SVM   &  0.91/0.06/0.10 &  69.2/0.57/10.5 &  64.7/0.67/12.2 &  60.0/0.37/0.70 &  41.2/0.40/0.74 &  41.4/10.1/16.2 \\
    LR    &  12.5/0.17/0.29 &  0.62/0.06/0.11 &  38.9/0.43/0.77 &  2.08/0.31/0.54 &  0.67/0.06/0.11 &  16.0/0.34/0.56 \\
    RF    &  38.2/11.7/17.9 &  54.1/20.9/30.1 &  52.5/19.0/27.9 &  52.1/22.8/31.8 &  45.6/17.9/25.7 &  35.9/11.8/17.7 \\
    DT    &  17.3/38.9/23.9 &  19.2/43.0/26.6 &  23.3/48.5/31.5 &  20.1/45.7/27.9 &  18.4/42.2/25.7 &  10.6/31.9/16.0 \\
    GBDT  &  26.1/28.9/27.4 &  30.3/27.8/29.0 &  38.9/30.1/33.9 &  33.6/24.7/28.5 &  31.3/27.2/29.1 &  25.3/21.0/22.9 \\
    XGB   &  34.5/22.8/27.4 &  41.2/26.6/32.3 &  50.0/27.6/35.6 &  \textbf{41.8/31.5/35.9} &  \textbf{47.3/30.6/37.2} &  \textbf{32.1/22.7/26.6} \\
    LGBM   &  \textbf{27.1/32.8/29.6} &  \textbf{30.2/38.6/33.9} &  \textbf{40.1/39.9/40.0} &  31.2/38.9/34.6 &  31.4/40.5/35.4 &  20.9/27.7/23.8 \\
    
    \hline
    \multicolumn{7}{c}{\textbf{Kernel\_Address}}\\
    \hline

    SVM   &  63.5/26.1/37.0 &  69.4/21.5/32.9 &  67.9/23.3/34.7 &  77.4/14.8/24.9 &  62.7/18.5/28.6 &  61.1/18.5/28.4 \\
    LR    &  64.7/24.4/35.5 &  70.6/22.8/34.4 &  53.2/20.2/29.3 &  56.5/16.0/25.0 &  53.8/16.2/24.9 &  25.3/18.5/21.4 \\
    RF    &  57.0/33.9/42.5 &  60.9/35.4/44.8 &  57.0/32.5/41.4 &  48.2/25.3/33.2 &  54.0/27.2/36.2 &  41.8/27.7/33.3 \\
    DT    &  22.6/42.2/29.4 &  25.9/48.7/33.8 &  24.7/50.3/33.1 &  21.9/47.5/30.0 &  21.4/44.5/28.9 &  12.3/44.5/19.2 \\
    GBDT  &  53.6/33.3/41.1 &  55.6/31.6/40.3 &  57.1/29.4/38.9 &  57.8/22.8/32.7 &  54.4/24.9/34.1 &  38.1/26.9/31.5 \\
    XGB   &  \textbf{58.0/36.1/44.5} &  \textbf{55.8/42.4/48.2} &  \textbf{58.9/40.5/48.}0 &  \textbf{58.5/34.0/43.0} &  \textbf{48.8/34.7/40.5} &  \textbf{37.6/34.5/36.0} \\
    LGBM   &  53.5/37.8/44.3 &  54.9/42.4/47.9 &  56.8/38.7/46.0 &  52.3/35.8/42.5 &  42.5/31.2/36.0 &  29.9/29.4/29.7 \\
    
    \bottomrule
    
    \end{tabular}}%
  \label{tab:clf}%
\end{table}%

DRAM prediction results are aggregated by serial number of servers, and only the first prediction result of each server every seven days is used. We compute Precision ($P$), recall ($R$), and F1-score ($F_1$) of failure servers to evaluate the performance of DRAM failure prediction. In terms of results of unsupervised anomaly detectors, since the returned result is an anomaly ranking rather than class labels, the area under the ROC curve (AUC) is commonly used in the literature \cite{liu2012isolation,pang2017hour,pang2021homophily,xu2018cikm,xu2019aaai,goldstein2012hbos,zhao2020copod,jian2018cure}. Following those prior work, AUC is used in evaluating the anomaly detection performance.

\subsection{Performance of Classifiers}

This experiment investigates the performance of seven commonly-used classifiers, including SVM with a RBF kernel (SVM), logistic regression (LR), random forest (RF), decision tree (DT), gradient boosting decision tree (GBDT), extreme gradient boosting (XGBoost), and light gradient boosting machine (LGBM), on the DRAM failure prediction datasets. The implementations of all these methods are taken from \texttt{sklearn} package, \texttt{xgboost} package, and \texttt{lightgbm} package. All of these classifiers are performed with their default settings. As original datasets are highly imbalanced, these classifiers are trained on the datasets pre-processed by a combination of over-sampling and under-sampling.


The precision, recall, and F1-score of different classifiers on these datasets are reported in Table \ref{tab:clf}. LGBM and XGB obtains more superior results than other classifier in most datasets. All classifiers perform poorly on the \textit{Address} and \textit{Mce} datasets, which indicates that more advanced models are required on those datasets.
A hybrid of over-sampling and under-sampling is used to overcome the issue of data imbalance, but more effective preprocessing methods, e.g., heuristic sampling or data augmentation, may be used for these datasets to yield more promising results. These classifiers can gain better F1-score in the two combined datasets -- \textit{Address\_Mce} and \textit{Kernel\_Address} -- than in single datasets. This may be because there are some important latent relationships between the features across these datasets.

\subsection{Performance of Anomaly Detectors}

As the whole data volume is very large, we test three representative fast anomaly detectors, including iForest\cite{liu2012isolation}, HBOS \cite{goldstein2012hbos}, and COPOD \cite{zhao2020copod}, in this experiment.

\begin{table}[htbp]
  \centering
  \caption{Six-fold Cross-Validation Results (AUC) of Different Anomaly Detectors on Five Datasets.}
    \scalebox{0.8}{
    \begin{tabular}{p{1.6cm}<{\centering}|p{1.6cm}<{\centering}|p{1.6cm}<{\centering}|p{1.6cm}<{\centering}|p{1.6cm}<{\centering}|p{1.6cm}<{\centering}|p{1.6cm}<{\centering}}
    \toprule
     \multicolumn{1}{c}{}    & \multicolumn{1}{c}{\textbf{Jan-Fold}} & \multicolumn{1}{c}{\textbf{Feb-Fold}} & \multicolumn{1}{c}{\textbf{Mar-Fold}} & \multicolumn{1}{c}{\textbf{Apr-Fold}} & \multicolumn{1}{c}{\textbf{May-Fold}} & \multicolumn{1}{c}{\textbf{Jul-Fold}} \\
    \midrule
  
    \multicolumn{7}{c}{\textbf{Kernel}}\\
    \hline
    
    iForest & 68.8  & 68.1  & 71.1  & 72.0  & 77.9  & 77.1  \\
    HBOS  & \textbf{75.1}  & \textbf{75.5} & \textbf{77.4}  & \textbf{87.1}  & \textbf{82.6}  & \textbf{79.5}  \\
    COPOD & 71.3  & 68.6  & 70.1  & 68.2  & 73.0  & 76.8  \\
    
    \hline
    \multicolumn{7}{c}{\textbf{Address}}\\
    \hline
    iForest & 80.4  & 78.7  & 80.9  & 75.9  & 78.9  & 78.2  \\
    HBOS  & 79.6  & 77.8  & 77.5  & \textbf{81.9}  & 80.0  & 77.8  \\
    COPOD & \textbf{81.0}  & \textbf{80.8}  & \textbf{81.6}  & 79.0  & \textbf{81.2}  & \textbf{78.4}  \\
    
    \hline
    \multicolumn{7}{c}{\textbf{Mce}}\\
    \hline
    iForest & 71.2  & 68.7  & 71.9  & 64.9  & 70.7  & 77.6  \\
    HBOS  & \textbf{77.7}  & \textbf{76.6}  & \textbf{77.2}  & \textbf{81.4}  & \textbf{79.9}  & \textbf{78.9}  \\
    COPOD & 35.0  & 40.9  & 40.1  & 42.0  & 38.9  & 54.3  \\
    
    \hline
    \multicolumn{7}{c}{\textbf{Address\_Mce}}\\
    \hline
    
    iForest & 79.3  & 76.7  & 79.0  & 72.5  & 77.6  & 79.3  \\
    HBOS  & 79.8  & 78.1  & 79.2  & \textbf{82.0}  & \textbf{80.6}  & \textbf{79.8}  \\
    COPOD & \textbf{80.5}  & \textbf{79.7}  & \textbf{80.5}  & 77.6  & 80.0  & 79.7  \\
    
    \hline
    \multicolumn{7}{c}{\textbf{Kernel\_Address}}\\
    \hline
    iForest & 67.1  & 65.5  & 71.6  & 70.2  & 76.3  & 76.0  \\
    HBOS  & \textbf{75.5}  & \textbf{76.4}  & \textbf{78.3}  & \textbf{86.0}  & \textbf{82.7}  & \textbf{79.9}  \\
    COPOD & 72.7  & 73.6  & 78.0  & 76.6  & 79.9  & 76.6  \\
    
    \bottomrule
    
    \end{tabular}}
  \label{tab:addlabel}%
\end{table}%

The AUC results of three anomaly detectors are reported in Table \ref{tab:addlabel}. HBOS is the best performer in most cases, while iForest is relatively less effective than the other two detectors. Specifically, HBOS outperforms iForest and COPOD in \textit{Kernel}, \textit{Mce}, and \textit{Kernel\_Address}. By contrast, COPOD obtains the best results in \textit{Address} and \textit{Address\_Mce}. 

Large-scale data centers may use DRAMs from various manufactures and vendors. It is therefore hard to guarantee that all types of installed DRAMs have labeled training data for supervised learning. Further, there can be unknown types of failures occurred in new DRAM models, to which classification methods fail to generalize. From these perspectives, unsupervised anomaly detection methods are more suitable than classification methods, as they are effective in detecting any DRAM failures, especially unknown failures, without requiring manually labeled data. 


\section{Challenges and Opportunities}

\subsection{Feature Engineering}
The provided dataset only contains raw features. Raw data is transferred to structured feature vectors via feature engineering so that machine learning algorithms can work on these feature vectors. Thus, how to effectively perform feature engineering becomes one of the key factors to success in this competition. High-quality features can bring significant improvement of failure prediction performance.

In other words, the success of the failure prediction solution depends mostly on the data representation rather than model selection and/or tuning. In many winning solutions of Kaggle and KDD Cup competitions, feature engineering also plays an important role, as shown in \cite{heation2017empirical}. An extreme situation is that even the simplest algorithm can obtain good results with the help of high-efficacy constructed features that are highly relevant to the expected prediction targets. 

In our feature engineering, we set a fixed time window (1 hour) and collect the summation and the number of past records in this time window as new features. It is also possible to use multiple time granularity to further enhance the performance. For example, the distribution of records in the past time windows can also be properly described in some new features. 

Some relevant studies in the area of anomaly detection create new features from the data itself. Hees et al. \cite{hees2021reconstruction} proposes a generic data preprocessing approach to generate additional features. For each original feature, a regressor is trained based on other features, and a new reconstruction error column is created based on the predicted value and the real value of each feature. This is similar to Zhao and Hryniewicki \cite{zhao2019xgbod} that uses anomaly scores derived by unsupervised anomaly detectors as new features. In this competition, Such new features can also be generated to better describe the data.

Instead of manually designing features, we can also achieve auto feature engineering by harnessing the strong representation power of deep learning. The usage of proper deep learning techniques can automatically generate dense low-dimensional and compressed representations. Although it is hard to know the semantic meaning of each dimension, this representation space is embedded with high-level semantics related to the task. Underlying relationships of the data records are encoded into this representation space. In this competition, there are still potential possibilities of using deep learning to capture more accurate and more informative temporal representation. For example, deep anomaly detection with Recurrent Neural Network (RNN), LSTM and GRU networks \cite{pang2021deep} may be applied to further enhance the performance.

\subsection{Categorical Features}
In the provided dataset, all the three tables contain categorical features. For example, Kernel table contains 24 binary features indicating whether the kernel log meets the given failure templates. The Mce table has ``mce\_id'' and ``transaction'' features. The Address table includes the discretized failure location. 

Feature values of these categorical features do not present explicit distance or similarity relationship. How to handle categorical data has been a challenging problem. Most of the notions like distance metrics, density, or projection that are popularly used in numerical space cannot directly applied in categorical space \cite{xu2019icdm}. Although one-hot encoding can transform categorical features into numerical data, 
the dimensionality of the transformed data increases to a large extent, and the transformed data is considerably sparse. These factors may pose significant challenges to these methods that are originally designed for common numerical data. 

There remains many opportunities of employing techniques that are specially designed for categorical data on the DRAM failure dataset. A number of categorical data-oriented methods are proposed for several tasks in data science, e.g., representation learning \cite{jian2018cure}, anomaly detection \cite{pang2021homophily,xu2019aaai}, and feature selection \cite{pang2017hour,xu2018cikm}. These techniques may be utilized to preprocess the DRAM failure datasets to obtain more expressive representations of their categorical features.

We present some possible methods of processing these categorical features here. Assume a dataset $\mathcal{X}$ be composed of $N$ data objects $\mathcal{X}=\{\mathbf{x}_1, \mathbf{x}_2, \cdots, \mathbf{x}_N\}$ described by a set of categorical features $\mathcal{F}=\{f_1, f_2, \cdots, f_D\}$. The value of feature $f$ in data object $\mathbf{x}$ is denoted as $v_f^{\mathbf{x}}$. 

Some notions in probability theory can be used to capture the abnormality of each individual feature value, and it is also possible to describe interactions and relationships between these categorical feature or feature values. We present marginal probability, joint probability, and Ochiai coefficient below. 

Marginal probability of categorical feature value $v$ can be computed as: 
\begin{equation}
    P(v) = \frac{|\{\mathbf{x}\in\mathcal{X} | v_f^{\mathbf{x}} = v\}|}{N}.
\end{equation}

Joint probability of feature value $v_i$ and $v_j$ is:
\begin{equation}
    P(v_i,v_j)=\frac{|\{{\mathbf{x}}\in\mathcal{X}| v_{f_i}^{\mathbf{x}} = v_i \cap v_{f_j}^{\mathbf{x}} = v_j \}|}{N}.
\end{equation}

Value similarity can be measured by using Ochiai coefficient:
\begin{equation}
    s(v_i, v_j)  = \frac{P(v_i, v_j)}{\sqrt{P(v_i)P(v+j)}}.
\end{equation}

Further, some notions in information theory, e.g., entropy and mutual information, can be derived. Entropy is defined as:
\begin{equation}
    H(\mathcal{F}) = \sum_{i=1}^{D} H(f_i | f_{i-1}, \cdots, f_1),
\end{equation}
while mutual information is defined as:
\begin{equation}
    I(f_i; f_j) = H(f_i) - H(f_i |f_j).
\end{equation}
These notions can describe feature-level information. For example, entropy indicates uncertainty relative to a random variable. A subset of DRAM records are abnormal if their removal from the whole dataset causes significant decrease of the entropy of the dataset \cite{wu2013information}. These notions are also relevant to assess the abnormality.

\subsection{Tree-based Methods vs. Neural Networks}

As practice shows, XGBoost and other gradient boosting models work better on these data mining competitions. However, this is not due to an inherent weakness of deep learning.

Theoretically, a neural network can be constructed to achieve the same performance as tree-based methods, but it is too hard to realize in practice for problems in data mining competitions because of limits of time and device. On one hand, deep learning achieves great performance with a appropriate structure of neural network, but the network architecture is hard to design for a given dataset in a short period of time. On the other hand, deep neural networks often require large-scale labeled data and extensive hyper-parameter (manual) tuning. By contrast, tree-based methods are relatively much easier to achieve better performance.

Data characteristics also determine the choice of approach. Neural network can perform better on the data that has homogeneous features, e.g., pixels in an image and frames in a video. However, data generated from feature engineering often contain heterogeneous columns in data mining competitions. In this competition, normalization methods is performed to scale each feature so that neural networks can work. By contrast, tree-based methods treat features independently, and they are able to build rules based on single features as well as the combinations of multiple features. Thus, tree-based methods are naturally more suitable for data with heterogeneous features.

\section{Conclusions}

In this paper, we discussed DRAM failure prediction from multiple aspects. In combination with our attempts in PAKDD 2021 Alibaba AIOps Competition, we conducted empirical evaluation to investigate the performance of seven commonly-used classifiers in the data of DRAM failures of large-scale data centers provided by Alibaba Cloud. Besides, three anomaly detectors are also used to explore the effect of unsupervised anomaly detection in this area. We also analyzed challenges and opportunities of the DRAM failure prediction task. This may foster future research on DRAM failure prediction techniques that can be used in AIOps of large-scale data centers.


\subsubsection*{Acknowledgements.} This work is supported by the National Key Research and Development Program of China (2016YFB1000101), the National Natural Science Foundation of China (No. 62002371, No.61379052, and No.61472439), the Science Foundation of Ministry of Education of China (No.2018A02002), and the Natural Science Foundation for
Distinguished Young Scholars of Hunan Province (No.14JJ1026).

\bibliographystyle{splncs04}
\bibliography{ref}

\end{document}